\documentclass[11pt]{article}

\usepackage[utf8]{inputenc}
\usepackage[T1]{fontenc}
\usepackage{microtype}

\usepackage[margin=1in]{geometry}

\usepackage{amsmath,amssymb}

\usepackage{graphicx}
\usepackage{float}

\usepackage{booktabs}
\usepackage{multirow}
\usepackage{array}

\usepackage[colorlinks=true,citecolor=blue,linkcolor=blue,urlcolor=blue]{hyperref}

\usepackage[numbers,sort&compress]{natbib}

\usepackage{titlesec}

\usepackage{caption}
\graphicspath{{figures/}}

\newcommand{\homeLocal}{home local server}
\newcommand{\localFirst}{local-first}

\title{\Large\bfseries A Generative AI-Driven Architecture for Whole-Home Intelligent Services: Edge Models, Memory Management, and Household Applications}

\author{%
  Mo Zhou\\
  \texttt{Nanjing Normal University}
}

\date{}

\begin{document}

\maketitle

\begin{abstract}
Generative artificial intelligence is entering the home. Cloud-centered conversational assistants face governance limitations in data transmission, continuous availability, and the reliability of physical actions. These limitations are particularly salient in households, where devices remain online, data accumulate over time, and commands can trigger real-world actions. Using the \homeLocal{} as a unifying organizing axis, this article develops a design-oriented review framework. Sources from arXiv, CNKI, ACL Anthology, and official platform documentation were searched in layers through 8 November 2026 and coded into five evidence categories: reviews, system prototypes, benchmark datasets, human-computer interaction and eldercare studies, and official documentation. System designs were then mapped onto a shared set of architectural modules. The review identifies nine recurrent responsibilities of the \homeLocal{}. Generative models support perceptual interpretation, intent understanding, and candidate planning, while deterministic validation, knowledge-graph constraints, and authorization leases confine candidate actions to an inspectable space. Home control, emotional companionship, and intelligent guarding can therefore be compared within a common framework: they share state and memory but differ in their requirements for latency, false positives and false negatives, proactivity, and permissions. Existing benchmarks rely on virtual or simulated environments and cannot substitute for evaluation in real deployments. Local deployment is not inherently safer, and technical continuity alone does not establish the social or medical effects of companionship and guarding. The article further proposes a five-dimensional evaluation framework covering architectural capability, system resources, privacy, safety and reliability, and human experience, together with six design principles and a future research agenda. It offers an auditable and layered governance reference for \localFirst{} intelligent home systems.
\end{abstract}

\vspace{1em}

\vspace{1em}

\section{Introduction: From Cloud-Based Smart Assistants to the Home Local Server}

Voice-driven smart assistants have become widespread in the consumer market, and the home is among the highest-frequency usage scenarios. Mainstream implementations upload speech and context to the cloud, where remote large models perform understanding and generation, and the results are returned to the terminal. This paradigm has low engineering cost for single-turn question answering, and model updates are rapid. The home scenario, however, differs from general information queries: devices remain online continuously, data accumulate over extended periods, commands can trigger real-world physical actions, and multiple members share the same dwelling. The problem therefore extends beyond one-time data transmission. Platforms can accumulate long-term behavioral profiles of users; once the network is interrupted, basic control fails; and after model updates, devices may exhibit behavioral changes that users did not anticipate.

Progress in edge models and home computing makes an alternative organizational approach feasible. Quantization and distillation enable language models at the billion-parameter scale to run on home gateways, small servers, or development boards, undertaking intent understanding, candidate planning, and dialogue generation. The home local server simultaneously hosts the device registry, state storage, rule engines, and audit logs, converging functions that were previously dispersed across the cloud and multiple vendor applications into a computational node within the dwelling. Here, local-first does not mean absolute disconnection from the network. The criterion adopted in this paper is ``local processing by default, with auditable cloud fallback under user authorization,'' which is explicitly distinct from the local-only mode of cutting all network connections.

The three task categories of control, companionship, and guarding are currently advanced in separate research communities. The control direction focuses on instruction understanding and device execution; the companionship direction focuses on persona and long-term memory; the guarding direction focuses on event recognition and risk escalation. The underlying problems they share are in fact the same: how to represent state, how to maintain memory, how to orchestrate models, and how to constrain actions. These problems have rarely been discussed under a unified architectural perspective, giving rise to concrete epistemic consequences. State representations or authorization mechanisms accumulated in one community cannot be borrowed by the other two; similar risks are repeatedly named in different applications without a common design vocabulary. Within the search scope and cutoff date set by this paper, no design-oriented review has been identified that uses the home local server as a unifying organizational axis while simultaneously integrating the three task categories. This judgment does not mean that no prior research exists in this direction; rather, it notes that existing surveys are bounded by edge-model technology or a single application category, lacking integration oriented toward home service architecture.

To address this gap, this paper undertakes three tasks. First, it establishes a unified architectural vocabulary, grouping the responsibilities of the home local server into nine modules. Second, it compares the module division and evidence of different systems, clarifying what the local node holds, what the cloud retains, and what constrains execution. Third, it derives a five-dimensional evaluation framework from the gaps left by existing benchmarks, and presents design principles and a research agenda.

\section{Conceptual Delimitation and Review Methodology}

\subsection{Core Concepts}

The whole-home intelligent service architecture referred to in this paper denotes the overall design that uses a computational node within the dwelling as a hub, organizing device connectivity, perception, state, memory, models, and execution into a continuously running system. The home local server is an umbrella term for forms including network-attached storage, home gateways, small x86 hosts, and edge devices; this paper focuses on its responsibilities and does not discuss specific hardware selection.

Four sets of distinctions affect subsequent judgments. The first is local-first versus fully offline. Local-first means that data and computation are completed within the dwelling by default, and cloud invocations require user authorization and leave traces; fully offline severs all network connections. The two must not be conflated. The second is home state versus home memory. Home state is a structured record of the device registry, room topology, current device states, and user preferences; home memory addresses longitudinal experiences, including dialogue history, user profiles, and long-term events, whose organizational scheme is elaborated in Section 4. The third is model versus Agent. Models provide reasoning capability; the home Agent, on top of the model, undertakes task decomposition, tool invocation, and result integration, serving as an execution unit. The fourth is generative planning versus deterministic execution. Generative models, with language generation at their core, assume responsibilities for perceptual interpretation, intent understanding, planning, and interaction; deterministic executors do not depend on generative output, relying on rules or type constraints to ensure that actions fall within the legal space. Edge models, as the carriers of the above roles, refer to models running on home or terminal hardware, including quantized and distilled small language models and embedded multimodal models.

The meaning of the term ``generative'' in the title must also be understood within these four sets of distinctions. Generative models assume responsibilities for interpretation and planning; yet non-generative modules---device connectivity, rule engines, deterministic verification, and auditing---are equally necessary components for the safe operation of the system. Prior studies have repeatedly shown that unconstrained open-ended generation produces invalid and even dangerous actions in physical control scenarios. The architectural stance of this paper is therefore the collaboration of these two classes of modules, rather than the replacement of all traditional components by generative models.

\subsection{Review Methodology}

The sources of this review comprise arXiv preprints, Chinese-language journals and dissertations indexed by CNKI, formally published entries in the ACL Anthology, peer-reviewed journal and conference papers, and official documentation of platforms such as Home Assistant. The search was cut off on 2026-11-08, and the arXiv literature mainly covers the years 2023 to 2026. Search terms were combined along three dimensions---``home/smart home'', ``local/edge/on-device'', and ``language model/Agent''---with the two thematic term groups ``memory'' and ``security and privacy'' added separately; supplementation was then performed by snowballing through the references of the included literature.

Inclusion followed three principles: direct relevance to the home scenario or to the responsibilities of the home local server; a verifiable formal source; and metadata confirmable from the publisher's page or the first page of the PDF. Retracted entries were excluded; platform documentation states only functionality and release dates and is not conflated with peer-reviewed evidence. Owing to the limitations of the existing search records, this paper does not report complete screening and exclusion counts for each stage, nor does it claim to have exhaustively covered the related research.

Evidence is differentiated across five levels. Review articles delimit the technical background and taxonomy; system prototypes carry architectural design and implementation details; benchmark datasets provide comparable evaluation results; human--computer interaction and eldercare research supply usage scenarios and user evidence; official platform documentation is used solely to state platform functionality and release dates. The core references are additionally subjected to structured field coding, with fields covering deployment location, generative role, non-generative modules, state, memory, orchestration, safety, cloud-edge collaboration, evaluation, and inter-module relations. Each system design is subsequently mapped onto a unified set of architectural modules, so as to compare how different systems divide the work of deployment, memory, orchestration, and safety.

The delimitation of the technical background rests mainly on review articles. A survey of edge large language models organizes the technical essentials of on-device deployment across three layers---model design, execution methods, and applications [1]. A survey of edge generative AI further consolidates the paths and constraints of migrating generative capabilities to edge devices [2]. A survey of smart spaces examines, as a whole, the methods for creating systems in indoor environments, which helps to delineate the relationship between the home scenario and smart spaces in a broader sense [3]. Chinese-language surveys provide a domestic scholarly lineage. A survey of on-device AI algorithms reviews the application status of edge models in vision, language processing, and the Internet of Things [4]; research progress on large models for edge intelligence summarizes technical challenges along the two threads of edge training and edge inference, and identifies high cost, limited scalability, and information security risks as the principal problems confronting the traditional cloud-based paradigm [5].

\section{Architectural Foundations of the Home Local Server}

\subsection{Partition into Nine Modules}

A cross-system comparison yields, first, one observation: the home local server repeatedly assumes an identical set of responsibilities; different systems merely name and combine them differently. This paper groups these responsibilities into nine modules, namely device connectivity and capability registration, perception and event extraction, home state, memory management, model serving, Agent orchestration, safe execution, cloud-edge collaboration, and feedback and auditing. 
The interoperability of the device connectivity layer determines whether the architecture can scale. Research built on the Matter standard combines a unified device standard with large-model applications, so that devices from different vendors can be invoked by models through a single semantic interface; this dissertation provides a complete engineering case from device connectivity to application implementation, indicating that standardized connectivity and upper-layer model orchestration require co-design [6]. Research based on edge computing decomposes an intelligent control framework into a layered structure of perception and connectivity, edge processing, and cloud collaboration, and examines its role in response speed and resource allocation through practical cases, which helps in understanding how the nine modules can be layered and realized in engineering practice [7].

\subsection{The Organizational Logic among Modules}

How modules are organized matters more than the module list itself. Judging from the commonality of existing systems, one recommendable pipeline is as follows: natural-language and multimodal models handle perceptual interpretation, intent understanding, and candidate planning; home state, memory, and the device capability registry provide structured context to the models; the policy layer handles model routing, user confirmation, and permission judgment; deterministic verifiers, graph constraints, or authorization leases confine candidate actions to a verifiable execution space; finally, device adapters carry out the actions and write the results into audit records.

Different systems arrange ``local'' in different ways, which can be roughly divided into three routes. The first route keeps models and profiles local. HomeLLaMA pairs a locally customized small model with a cloud-based teacher model, organizing local user profiles and device discovery into a privacy-preserving service path, with emphasis on the coordination between personalization and local profiles [8]. Harmony combines a locally deployable model with modular Agents and structured prompts, controlling hallucination and misinterpretation through module division of labor and prompt constraints [9]. The second route keeps state, orchestration, and authorization on edge nodes, while inference may be offloaded. HearthNet implements multi-Agent orchestration and Git-managed shared state on edge hardware, coordinating planning, verification, authorization, and execution through a message bus, execution leases, and timeline tracking; its inference stage uses hosted interfaces, taking the form of ``orchestration and state local, inference offloadable'' rather than that of a fully local system [10]. The third route embeds models directly into devices or IoT networks. Thoughtful Things places small language models into individual devices, completing interaction and interpretation locally [12]; Talk with the Things integrates language models into IoT networks, so that devices are invoked in a unified manner [13]. The three routes jointly indicate that the crux of local-first is not that all inference runs offline, but rather which state and control stages the home node holds.

All three routes involve the boundary between generation and execution. The dissertation \textit{Speaking to the Edge} installs deterministic gating between conversational control and physical execution, verifying model-generated control instructions before driving devices [11]. This gating mechanism can be combined with the three deployment routes described above; its function is to confine open-ended generation within a verifiable action space.

\begin{figure}[H]
  \centering
  \includegraphics[width=\textwidth]{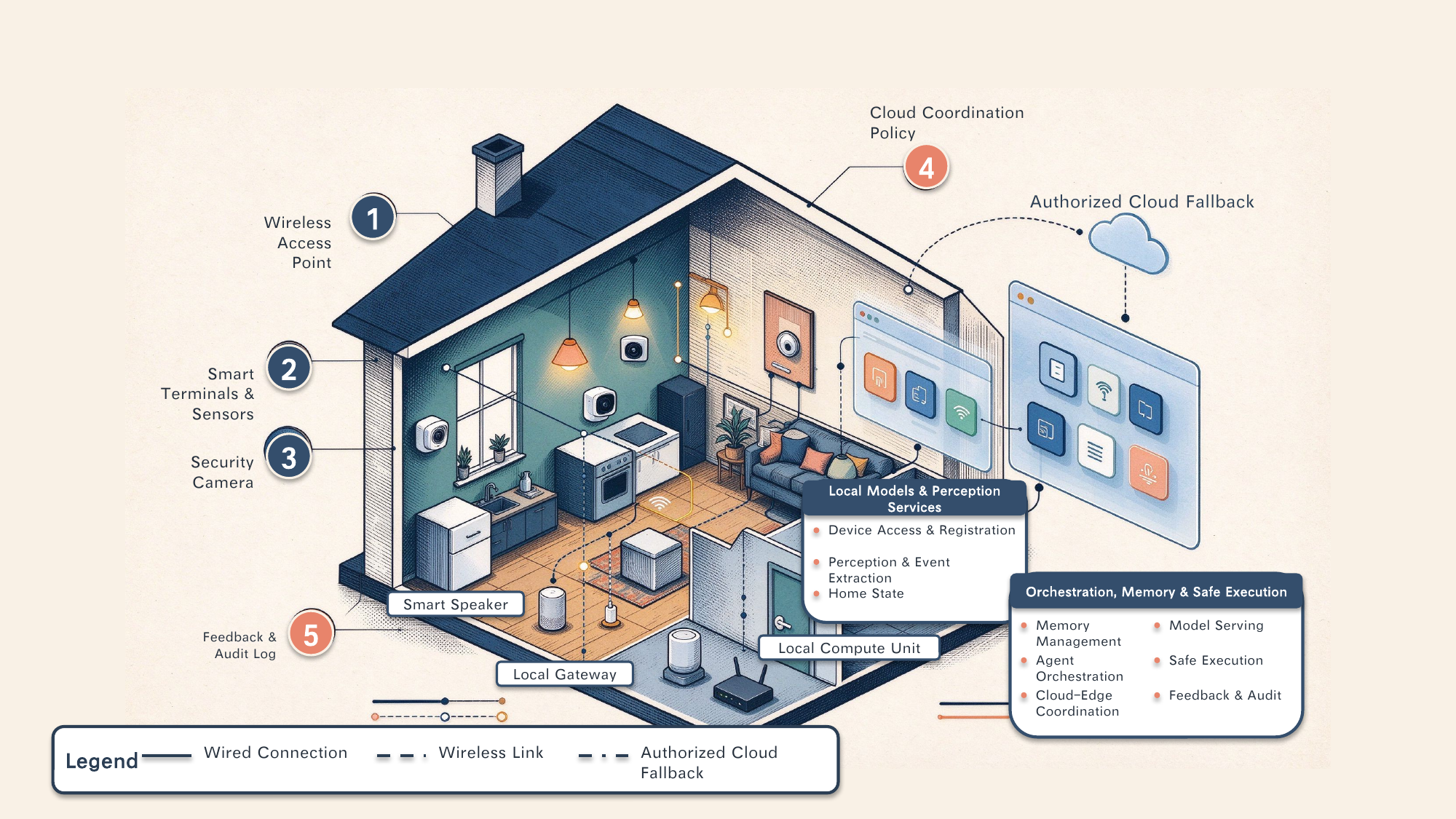}
  \caption{Overall architecture of the home local server.}
  \label{fig:architecture}
\end{figure}

The actual coverage of individual systems varies, and no single paper proposes a complete nine-module architecture: some focus on models and profiles, some on orchestration and authorization, and some on single-device embedding. The significance of mapping them onto unified modules lies in identifying which responsibilities recur in home services, which stages already possess relatively mature engineering patterns, and which modules still lack reliable implementations. The nine modules simultaneously serve as the analytical language of the subsequent chapters: Section 4 focuses on state, memory, and model serving; Section 5 links orchestration and execution through the thread of applications; Section 6 concentrates on safe execution and cloud-edge collaboration.

Home state must both provide context to models and constrain the action space; memory management needs to cover writing, maintenance, retrieval, and deletion; and model orchestration allocates capabilities according to call sites. Together, these three determine whether a home local server can remain controllable under continuous operation.

\section{Home State, Memory, and Model Orchestration}

\subsection{Organization of Home State}

Home state provides the factual basis for all upper-layer functions and consists of the device registry, room topology, current device states, user preferences, and daily patterns. It is not merely a context for models to read; it is also a constraint that limits the action space. Trail constrains the candidate space with a runtime registry, replacing open-ended generation with typed decisions oriented toward rooms, devices, and operations, and employs a confidence mechanism to decline uncertain instructions rather than forcibly executing them \cite{masood2026trail}. This design clearly shows that the registry is, first and foremost, a structural constraint on the model's output space and only secondarily an inventory of devices.

The representation of preferences and daily patterns determines whether the system can handle personalized instructions. AdaHome dynamically selects between a simple path and a reasoning path through intent-aware routing, and updates preferences via user feedback without retraining or prompt expansion; it supports a design judgment: routing and preference adaptation matter more than choosing a single strongest model \cite{lim2026adahome}. Related studies also show that context organization combining the home environment with user history can improve response relevance, but the effect depends on the degree of structuring of the context rather than its simple accumulation \cite{rey2024leveraging}.

\subsection{Layering and Maintenance of Memory Management}

Short-term context and long-term memory require layered organization. The former serves multi-turn dialogue and the resolution of elliptical expressions; the latter includes user profiles, event experiences, and preference evolution. PEC-Home, targeting progressive elliptical instructions, separately examines the roles of dialogue history tools and environment and preference context, finding that retrieving dialogue history alone is insufficient to correctly resolve elliptical expressions and that environmental state and user preferences are equally indispensable \cite{shan2026pechome}.

Long-term memory should be understood through a lifecycle of ``writing, maintenance, retrieval, and deletion or governance,'' rather than through retrieval alone. Embedded AI Companion distinguishes active and idle periods on edge devices: retrieval supports real-time dialogue during active periods, while more costly memory extraction, consolidation, and maintenance are performed during idle periods, explicitly framing memory scheduling as a server-level resource and latency design problem \cite{gupta2026embedded}. Mnemosyne stores long-term memory in a graph structure, accompanied by lifecycle operations such as write filtering, deduplication, commit and pruning, and temporal decay and refresh, and evaluates maintenance mechanisms beyond retrieval on longitudinal dialogue and temporal reasoning benchmarks \cite{jonelagadda2025mnemosyne}. Different households differ in write frequency, storage budget, and privacy constraints; no single memory architecture can serve as a universal solution, and memory strategies must be matched to local resource conditions.

The evaluation of retrieval augmentation must be bounded within specific experimental settings. The observation that ``retrieving dialogue history alone is insufficient for resolution'' indicates that vector retrieval cannot replace environmental state and preference modeling; however, this conclusion applies only to the tasks and tool configurations of that benchmark and cannot be generalized to the ineffectiveness of retrieval-augmented methods in general. The design of a memory system must therefore answer three interrelated questions: what content is worth writing, how to deduplicate and prune after writing, and how to combine temporal and situational information at retrieval. If any link is missing, memory can be ``stored but not well used.''

\subsection{Model Orchestration}

Model orchestration organizes model families by responsibility rather than by model size. What a system needs to determine is which capability to invoke under which circumstances: when to read memory, when to enable multimodal perception, when to require human confirmation, and when to trigger cloud fallback. Not Every Call Needs a Frontier Model evaluates the suitability of small models at each call site and finds that small models suffice at some call sites while stronger capabilities are required at others, providing empirical grounds for allocating models by call site \cite{kasnesis2026not}.

Collaborative inference and training mechanisms further expand the orchestration space. A survey of collaborative inference between small edge models and large cloud models classifies existing algorithms by execution mode and division of labor, showing that local and remote models can collaborate in segments within the same task pipeline \cite{li2025collaborative}. On the training side, Edge-FIT performs instruction fine-tuning of quantized models in a federated manner, enabling home environment data to participate in model adaptation without being centrally exported \cite{venkatesh2025edgefit}. Orchestration is therefore not a one-time model selection but a continuous decision encompassing inference division of labor, training adaptation, and fallback governance.

\section{Home Applications: Home Control, Emotional Companionship, and Intelligent Guarding}

Home control, emotional companionship, and intelligent guarding share home state and memory, but differ in their requirements for module composition, evidence types, and risk boundaries. Comparing these three application categories tests whether a unified architecture suffices to cover divergent task demands.

\subsection{Home Control and Natural Human-Computer Interaction}

The evolution of home control capabilities exhibits a clear progression: from explicit single-device commands, to multi-device coordination, and then to vague and elliptical expressions, preference dependence, persistent rules, and failure recovery with safe refusals. Early work established the basic paradigm of goal inference and grounded execution. Sasha targets creative goal-directed reasoning, decomposing vague goals into executable home tasks \cite{king2023sasha}; SAGE employs a grounded execution mechanism to ensure that device actions invoked by the Agent correspond to real capabilities \cite{rivkin2023sage}.

On resource-constrained local hardware, there is direct experimental evidence on whether a single model can undertake both understanding and generation. On-Device LLMs for Home Assistant assigns a single model both intent slot detection and response generation, outputting JSON action calls and natural language. Evaluation shows that the 16-bit and 8-bit quantized versions maintain high accuracy, while the 4-bit version degrades noticeably on device service classification; accuracy on noisy and out-of-domain instructions is approximately 80--86\%, with an average inference latency of 5--6 seconds, suitable for one-shot commands but unfavorable for multi-turn dialogue \cite{birkmose2025ondevice}. These results simultaneously demonstrate the feasibility of the local control interface and the trade-off between latency and quantization.

Control performance depends on the combination of the model with the device registry, home state, memory, and a deterministic validator, rather than on the model alone. The model handles understanding and candidate planning; the executor confines actual actions to the legal space. HomeBench supports this judgment from the opposite direction: by testing 13 models with valid and invalid commands and single-device and multi-device tasks, it finds that invalid multi-device commands are a pervasive failure hotspot, and that architecture must treat validity checking and multi-device execution verification as first-class responsibilities \cite{li2025homebench}.

Chinese industrial practice reveals the direction of evolution for interaction entry points. Research on smart-speaker voice interaction proposes a technical path combining a full-duplex dialogue engine, role-playing agents, and voice cloning, shifting speakers from single-turn command responses toward interruptible, continuous natural dialogue, and connecting external knowledge and device control through tool-invocation mechanisms \cite{zhao2026smartbox}. Interaction upgrades at entry devices must be matched by orchestration and execution capabilities on the home server side; improving the dialogue experience at the entry point alone cannot guarantee that physical actions are reliably carried out.

\subsection{Emotional Companionship: Local Memory and Sustained Relationships}

Emotional companionship relies on four elements: local persona, long-term memory, proactive interaction, and emotional safety. Systematic research on persona design organizes the personas of companionship applications into a four-quadrant technical taxonomy, providing a structured language for persona selection across scenarios \cite{sun2025persona}. The aforementioned dual-phase maintenance of active and idle periods and the graph-structured memory architecture enable companionship systems to maintain longitudinal continuity under resource constraints \cite{chollet2024evita}.

Identity recognition and privacy boundaries in multi-member households are unavoidable. e-VITA implements speaker diarization and spoken dialogue on the device, allowing the conversations and profiles of different members to be distinguished and processed locally \cite{pedersen2025empathetic}. Research on empathetic robot design for elderly users summarizes expression and interaction strategies in age-friendly scenarios, emphasizing that persona settings must match user characteristics \cite{hu2026pattern}.

Proactive interaction is the key capability that distinguishes companionship from passive response, and it is also the most prone to overstepping. Local profiles and memory enable the system to initiate reminders or conversations at appropriate moments; however, if the frequency of proactivity, the range of topics, and quiet hours lack user-controllable policies, continuous care becomes persistent disturbance. Trigger rules should be made explicit and allow users to adjust or even disable them by scenario, consistent with the overall approach of tiered permissions.

Maintaining continuous dialogue and personalized responses at the technical level does not mean that effectiveness at the social level has been established. This paper does not describe companionship systems as means to ``improve mental health'' or ``alleviate loneliness''; the relevant longitudinal psychological effects require support from independent human-computer interaction research designs. A systematic review of mental health scenarios shows that model applications in this field remain at the stage of pattern exploration, with limited evidential strength \cite{mihalcea2026model}. Research on local-first psychological counseling architectures proposes the judgment that ``the model is not the product,'' emphasizing that deterministic pillars and generative pillars require side-by-side design \cite{choi2025desamo}.

\subsection{Intelligent Guarding: From Perceptual Triage to Risk Escalation}

The pipeline of intelligent guarding can be summarized as: perceptual input, local event recognition, integration of state and behavioral history, anomaly judgment, and risk grading, followed by escalation to personal reminders, family notifications, human care, or emergency services. Perception and event extraction research first addresses ``what has been detected.'' DESAMO processes raw audio directly with an embedded audio model, resolving ordinary control intents and critical events such as falls and calls for help uniformly at the same node, avoiding error accumulation caused by multi-level speech recognition cascades \cite{sun2024activity}. Such single-node schemes shorten the recognition chain, but their evidential scope remains limited to established sensing modalities and event categories and cannot independently demonstrate that the system can continuously detect deviations in household behavioral patterns.

Environmental sensor and multimodal studies supplement the event context. Sensor events are first transformed into activity sequences, which are then interpreted by a language model in context, providing a prototypical path for complex activity tracking and behavioral pattern analysis \cite{jung2026omni}. OmniSmartHome incorporates speaker identity, spatial location, and audio-visual cues into a multimodal evaluation, whose baseline Agent coordinates the various perceptual tools with procedural memory \cite{khalil2025redefining}. The former demonstrates a system implementation approach, while the latter provides a tool for comparing model reasoning capabilities; the evidence of both concentrates mainly on recognition and reasoning performance and cannot as yet substitute for long-term deployment testing in real homes.

Demand-side and service research addresses ``how to act after recognition.'' Research on agents for elder-care scenarios surveys the opportunities and challenges of care tasks, providing scenario grounds for risk grading and human handover \cite{huang2025community}. Research on community-based home care service systems discusses platform combinations of safety protection, personalized diet, and companionship \cite{yin2025elderly}; a voice assistant for elderly companionship implements dialogue and reminders locally \cite{marghalani2026shacr}. These studies provide demand and functional evidence rather than recognition accuracy evidence. Synthesizing the three categories of material, a guarding system needs to connect perceptual triage, contextual interpretation, and human escalation into a chain; event recognition constitutes only a risk indication, not a substitute for medical diagnosis or professional care, and prototype results cannot be extrapolated as long-term effects in real homes.

\subsection{Cross-Application Comparison}

The three application categories share home state, memory, registry constraints, and audit records, but impose different parameter requirements on the same set of modules. Home control is latency-sensitive and requires understanding, verification, and execution to be tightly coupled; its failures manifest primarily as erroneous execution. Emotional companionship imposes higher demands on proactivity and persona consistency, with risks lying in intrusive overreach and privacy confusion. Intelligent guarding must balance false alarms against missed detections simultaneously, impose stricter constraints on missed detections and escalation latency, and retain a human escalation path, so that triage conclusions do not directly become final judgments. The role of the unified architecture is to let shared modules adjust their configurations by task, not to equate the requirements of the three application categories.

\section{Security, Privacy, and Cloud-Edge Collaboration}

The safety constraints on home actions can be distinguished by point of action into pre-execution, in-execution, and post-execution. Local deployment does not inherently confer safety; local systems may likewise suffer from configuration errors, permission abuse, and tool-chain attacks. Safety depends on concrete mechanisms rather than on deployment location itself.

\subsection{Local-by-Default and Tiered Authorization}

The risks of home actions are not uniform, and permission design should reflect a gradient. From low to high risk, actions can be divided into ordinary queries, routine controls such as lighting and air conditioning, camera analysis, door lock operations, and high-risk alerts. Low-risk actions can be executed automatically on-premises; the closer an operation is to physical safety and privacy sensitivity, the stronger the confirmation and authorization required. The execution lease mechanism of HearthNet embodies this approach: the Agent may drive execution only upon obtaining a time-limited authorization, and expired or unauthorized commands are rejected \cite{zhan2025aegismcp}.

\subsection{Structured Safety Mechanisms}

Pre-execution constraints first address rule conflicts. SHACR represents devices, capabilities, states, and topology as a typed knowledge graph, confines the large model's interpretations and repair suggestions within the graph boundaries, and completes validation before execution through a closed loop of ``scan, explain, repair, and verify.'' Its testing covers 203 rules and 70 apartment configurations, providing experimental evidence for semi-autonomous conflict resolution under graph constraints \cite{zhan2026attack}.

Post-execution monitoring and auditing primarily target the tool chain. AegisMCP records behaviors among Agents, tools, devices, remote services, and sessions as temporal graphs, performs online anomaly detection on streaming events, and can identify patterns such as privilege escalation, data exfiltration, malicious registration, and persistence; detection completes at sub-second latency and is suitable for CPU edge hardware \cite{wang2026smarthomesecure}. System-level attack surface research further points out that the security risks of edge Agent deployment are distributed across the tool, session, and device layers, and that the provenance and permissions of the tool chain should themselves be treated as observable security state \cite{khatiwada2025llmiot}. Beyond the execution chain, configuration maintenance is also indispensable. SmartHomeSecure performs syntactic and common semantic error detection on Home Assistant YAML configuration files and completes automated repair through deterministic repair and constrained generation, complementing efforts to reduce automation failures caused by configuration errors \cite{schoutsen2024aiagents}.

\begin{figure}[H]
  \centering
  \includegraphics[width=\textwidth]{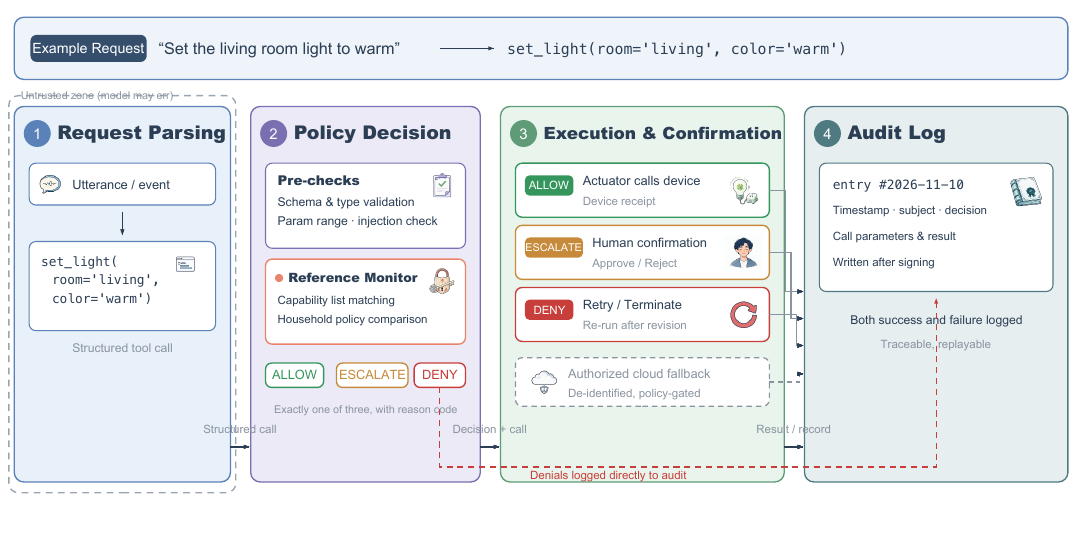}
  \caption{The safe execution chain: from events to validation, execution, and alerting.}
  \label{fig:safety_chain}
\end{figure}

\subsection{Privacy and Auditable Cloud Fallback}

A security survey of language models in IoT ecosystems shows that threat types span the data, model, and application layers, and that deployment location cannot substitute for mechanism design \cite{schoutsen2025localai}. In designs that retain cloud fallback, auditability is central to privacy governance. Every remote call should record the triggering reason, the fields transmitted, the service used, the returned result, whether execution occurred, and the user's confirmation status, rendering fallback behavior visible and traceable to the user.

Platform-side practice offers a reference. Home Assistant describes, in its official updates, designs including selectable local and cloud models, the exclusion of generative models from critical tasks, and per-device model assignment \cite{li2026devicesworld}. These materials are used solely to describe platform functionality and do not constitute peer-reviewed conclusions regarding security.

\section{Unified Evaluation Framework, Design Principles, and Future Agenda}

\subsection{Positioning and Limitations of Existing Benchmarks}

Existing benchmarks respectively examine instruction validity, context resolution, multimodal reasoning, and cross-device execution, covering different facets of architectural capability. HomeBench focuses on valid and invalid command handling \cite{li2025homebench}, and PEC-Home focuses on elliptical expressions and their dependence on context and preferences \cite{shan2026pechome}; the two indicate that validity checking, home state, and preference context should enter basic capability evaluations. OmniSmartHome examines multimodal reasoning with 1,360 synthetic event clips and 272 real event clips \cite{jung2026omni}, while DevicesWorld targets cross-device Agent evaluation in heterogeneous environments \cite{li2026devicesworld}; the two broaden perceptual cues and device environments, but do not yet cover long-term memory maintenance, permission escalation, and cloud fallback auditing.


\begin{table}[H]
\centering
\small
\caption{Selected classification of core literature by evidence level and architectural responsibility}
\label{tab:lit_classification}
\begin{tabular}{@{}p{3.8cm}p{2.2cm}p{4.5cm}p{2.5cm}@{}}
\toprule
\textbf{Source} & \textbf{Evidence Level} & \textbf{Primary Architectural Responsibilities} & \textbf{Environment} \\
\midrule
HomeLLaMA & System prototype & Model serving, memory, cloud-edge collaboration & Local + optional remote \\
Harmony & System prototype & Model serving, Agent orchestration & Local virtual environment \\
On-Device LLMs for HA & System prototype & Model serving, safe execution & CPU edge hardware \\
AdaHome & System prototype & Agent orchestration, preference adaptation & Local evaluation \\
Trail & System prototype & Typed decisions, abstention & Fully local benchmark \\
DESAMO & Prototype device & Audio perception, event recognition & Embedded device \\
HearthNet & System prototype & Multi-Agent orchestration, authorization leases & Edge + hosted inference \\
Speaking to the Edge & PhD thesis & Deterministic gatekeeping, device control & Local integrated system \\
Embedded AI Companion & System prototype & Biphase memory maintenance & Edge device \\
Mnemosyne & Architecture study & Long-term memory lifecycle & Edge-compatible architecture \\
SHACR & System prototype & Graph constraints, conflict resolution & Rule testbed \\
AegisMCP & Security framework & Tool-chain temporal monitoring & Edge testbed \\
HomeBench & Benchmark & Invalid instruction evaluation & Simulated environment \\
PEC-Home & Benchmark & Elliptical command resolution & Simulated home \\
OmniSmartHome & Benchmark & Multimodal reasoning evaluation & Synthetic + real episodes \\
DevicesWorld & Benchmark & Cross-device Agent evaluation & Heterogeneous simulation \\
e-VITA & System prototype & Speaker diarization, dialogue & Local device \\
Edge-FIT & Training method & Federated instruction tuning & Privacy-preserving environment \\
\bottomrule
\end{tabular}
\end{table}

\begin{table}[H]
\centering
\small
\caption{Responsibilities, primary risks, and recommended evaluation metrics for the nine modules}
\label{tab:nine_modules}
\begin{tabular}{@{}p{3.2cm}p{3.8cm}p{3.8cm}p{3.8cm}@{}}
\toprule
\textbf{Module} & \textbf{Core Responsibility} & \textbf{Primary Risk} & \textbf{Recommended Metrics} \\
\midrule
Device Access \& Capability Registry & Multi-protocol device management and capability registration & Capability misdescription, registry hijacking & Access scale, registration failure rate \\
Perception \& Event Extraction & Signal and multimodal input to structured events & Event misextraction, missed detection & Event accuracy, miss rate \\
Home State & Registry, topology, and current state maintenance & State inconsistency, staleness & State consistency, update latency \\
Memory Management & Layered short-term context and long-term memory maintenance & Memory contamination, sensitive accumulation & Retrieval accuracy, pruning coverage \\
Model Serving & Local model running, quantization, scheduling & Quantization degradation, resource contention & TTFT, memory footprint \\
Agent Orchestration & Task decomposition, tool calls, result integration & Tool misuse, runaway loops & Task completion, tool compliance rate \\
Safe Execution & Candidate action validation and legal-space constraints & Unauthorized execution, rule conflicts & Unauthorized action rate, refusal rate \\
Cloud-Edge Collaboration & Authorized remote calls and fallback & Excessive data outflow, invisible fallback & Outbound volume, fallback audit completeness \\
Feedback \& Auditing & Key event recording and retrospective support & Log missing, tampering & Audit completeness, traceability coverage \\
\bottomrule
\end{tabular}
\end{table}


\begin{figure}[H]
  \centering
  \includegraphics[width=\textwidth]{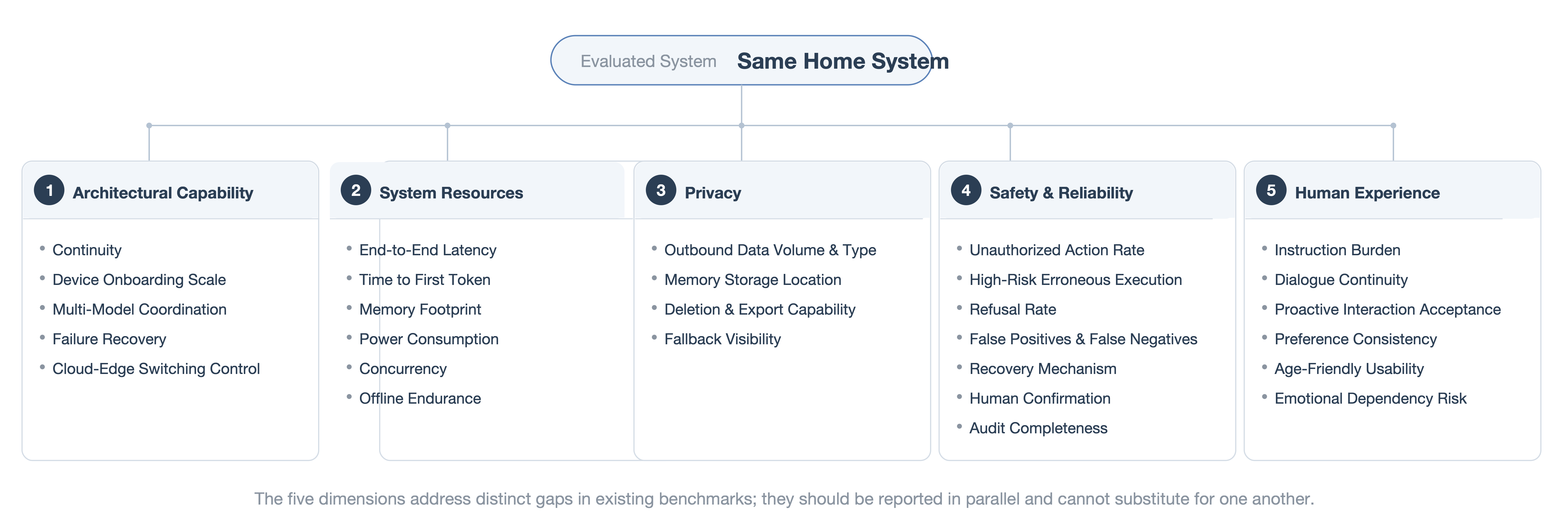}
  \caption{Five-dimensional evaluation framework for \localFirst{} intelligent home systems.}
  \label{fig:eval_framework}
\end{figure}

\subsection{Design Principles}

The evidence presented above can be compressed into six design principles. First, separate generation from execution: models produce candidates, while deterministic modules handle verification and execution. Second, state precedes intelligence: constrain the model's output space with the registry, topology, and state. Third, layer memory and maintain its lifecycle: configure write filtering, pruning, and temporal policies. Fourth, orchestrate models by responsibility: allocate capabilities by call site and retain auditable cloud fallback. Fifth, tier permissions by risk: the closer an action is to physical safety and sensitive data, the stronger the confirmation required. Sixth, ensure end-to-end auditability: record key events, remote invocations, and human confirmations to support post-hoc traceability.

\subsection{Future Research Agenda}

The future research agenda corresponds item by item to the gaps identified above. Regarding the disconnect between simulated environments and real deployments, behavioral drift caused by model updates lacks mature governance mechanisms, requiring version management and regression testing methods oriented toward home scenarios. Regarding cloud fallback and data minimization, operational rules need to be established between functional availability and data minimization. Regarding memory accumulation and deletion governance, the memory store itself may become a concentration point for sensitive data, requiring dedicated design for encryption, access control, and deletion mechanisms. Regarding the lack of unified standards for age-friendly evaluation, existing evidence largely comes from prototypes and small-scale settings, and more real-environment studies are needed. The balance between privacy and usability in multimodal perception, and the practical feasibility of federated and collaborative training under home bandwidth conditions, also await further experimentation.

\subsection{Conclusion}

Using the home local server as an organizing axis, this paper synthesizes nine modules into a unified analytical framework and surveys the evidence across three application categories on this basis. The central judgment is this: the key to local-first intelligent homes does not lie in shrinking a cloud-based large model and deploying it on a home server, but in organizing state, memory, model routing, permission policies, and deterministic execution so that home intelligence becomes a continuously running, auditable, and layered-governance computing system. What this paper offers is an architectural framework and design possibilities, not a finished system validated in real homes. Local deployment is not inherently safer, and the social and medical effects of companionship and guarding do not automatically follow from technical continuity. The value of subsequent research will be tested through the completion of real-deployment evidence and governance mechanisms.

\bibliographystyle{plainnat}
\bibliography{references}

\end{document}